\documentclass[10pt,twocolumn,letterpaper]{article}

\usepackage[pagenumbers]{cvpr} 

\usepackage{cuted} 

\usepackage{currfile} 

\usepackage{caption} 
\definecolor{cvprblue}{rgb}{0.21,0.49,0.74}
\usepackage[pagebackref,breaklinks,colorlinks,allcolors=cvprblue]{hyperref}

\def\paperID{*****} 
\def\confName{CVPR}
\def\confYear{2026}

\title{TemPose-TF-ASF: Two-Stage Bidirectional Stroke Context Fusion for Badminton Stroke Classification}

\author{Tzu-Yu Liu\\
Institute of Communications Engineering\\
National Tsing Hua University, Hsinchu City, Taiwan\\
{\tt\small 113064514@office365.nthu.edu.tw}
\and
Duan-Shin Lee\\
Department of Computer Science\\
National Tsing Hua University, Hsinchu City, Taiwan\\
{\tt\small lds@cs.nthu.edu.tw}
}

\begin{document}
\maketitle
\begin{abstract}
Accurate badminton stroke prediction is crucial for fine-grained sports analysis and tactical decision support.
 However, existing methods struggle to model rich temporal context.
 This paper introduces TemPose-TF-ASF (Adjacent-Stroke Fusion), a context-aware extension of TemPose.
 It enhances stroke recognition by incorporating stroke-type information from both preceding and subsequent strokes.
 A two-stage training and inference strategy is adopted.
 Preliminary predictions from the baseline model are reused as estimated temporal context.
 These predictions guide the joint optimization of the ASF module and the classifier.
 By explicitly modeling bidirectional temporal stroke dependencies, the proposed method can be seamlessly integrated into existing state-of-the-art models.
 Experiments on a large-scale badminton match dataset show consistent improvements over the baseline and its variants in terms of Accuracy and Macro-F1.
 Moreover, integrating ASF into other advanced methods yields notable performance gains.
 These results demonstrate strong transferability and generalization capability.

\end{abstract}
    
\section{Introduction}
\label{sec:intro}
With the rapid advancement of computer vision and deep learning technologies, sports action recognition has attracted increasing attention for applications such as match analysis, tactical support, and sports training \cite{BroadcastBadmintonAnalysis,BasketballValidateOpenPose,BasketballActionPrediction,SoccerMap,BaseballDL}. Among various sports, badminton stroke actions are characterized by a fast tempo and strong temporal dependencies, making accurate stroke-type recognition particularly challenging.
In recent years, numerous studies have sought to improve stroke recognition performance by leveraging skeleton sequences, shuttlecock trajectories, or multimodal features in conjunction with temporal models \cite{chang2025bst,TemPose,taichi,TableTennis,action,MonoTrack}. However, most existing methods focus on predicting the current stroke segment and overlook bidirectional temporal relationships in real matches.
In practice, stroke selection is influenced by both preceding strokes and upcoming tactical transitions.
As a result, relying solely on the current segment limits a model’s ability to capture the full semantic context of stroke actions.
To overcome these limitations, the TemPose-TF (Temporal Fusion) framework \cite{TemPose} is extended, and the TemPose-TF-ASF (Adjacent-Stroke Fusion) model is introduced. TemPose-TF is a skeleton-based Transformer architecture specifically designed for fine-grained badminton stroke recognition. A factorized Transformer encoder separately models the temporal dynamics of human motion and inter-player interactions.
Multimodal information, including skeletal joints, player court positions, and shuttlecock locations, is fused at an early stage.
This architecture has shown strong performance in badminton stroke recognition.
Building on this foundation, TemPose-TF-ASF integrates stroke-type information from both preceding and subsequent strokes to enhance bidirectional temporal context modeling.
Instead of using ground-truth annotations of future strokes, we introduce a Two-Stage Contextual Refinement (TSCR) training and inference strategy.
In the first stage, the baseline model generates preliminary predictions.
These predictions are then treated as estimated contextual information from adjacent strokes and used to jointly optimize the ASF module and the classifier.
This design enables effective use of temporal semantic context during inference while reducing the training–inference mismatch.
The main contributions of this work are summarized as follows:
(1) TemPose-TF-ASF is introduced to enhance temporal context modeling for badminton stroke recognition by integrating semantic information from both preceding and subsequent strokes.
(2) A Two-Stage Contextual Refinement (TSCR) strategy is proposed to mitigate the training–inference mismatch without relying on ground-truth future stroke annotations.
(3) Extensive experiments on the large-scale ShuttleSet dataset~\cite{ShuttleSet} show that the proposed approach significantly outperforms the baseline and its variants in Accuracy and Macro-F1.
(4) The ASF module is shown to be transferable, as it can be integrated into multiple state-of-the-art (SOTA) temporal action recognition methods~\cite{chang2025bst,taichi,SkateFormer,TemPose,BlockGCN}, consistently improving performance.
\section{Related Work}
\label{sec:related_work}
\subsection{Badminton Stroke Recognition and Multimodal Temporal Modeling}
Action recognition in racket sports is challenging due to rapid movements, visually similar strokes, and strong temporal dependencies. Early methods relied on handcrafted features and classical classifiers, such as HMMs \cite{retracted, BARRGB}, which are sensitive to environmental changes. CNNs later enabled learning discriminative spatial features \cite{CNN, AlexNetCNN}, but without explicit temporal modeling, performance remained limited.  
Recent approaches integrate CNNs or Vision Transformers with temporal models, including RNNs, LSTMs \cite{LSTM}, or Transformers \cite{Transformer,chang2025bst,CLSTM,TemPose}, to capture short- and long-term dependencies. Multimodal strategies further enhance recognition by fusing skeleton keypoints, shuttle trajectories, and RGB features, as in TemPose \cite{TemPose} and BST \cite{chang2025bst}, improving robustness and fine-grained stroke classification in complex match scenarios.

\subsection{Sequential Dependency and Contextual Refinement}
Modeling sequential dependencies is critical for handling rapid, temporally correlated strokes. Insufficient temporal context often leads to misclassification of visually similar or transitional actions. Multi-stage refinement methods, such as MS-TCN \cite{MS-TCN}, iteratively correct predictions using expanded temporal context, while bidirectional models \cite{BiLSTM} exploit past and future information to improve discrimination. Integrating skeletal motion, object trajectories, and temporal modeling has been shown to enhance sequence-level consistency and recognition accuracy \cite{ShuttleNet, Timerefine, ContextRefinement}.

\section{Method}
\label{sec:method}
\subsection{Deriving Player Motion and Stroke Context from Video Clips}
\label{subsec:data}
TemPose~\cite{TemPose} proposes an intuitively structured and efficient skeleton-based Transformer that models individual temporal dynamics and inter-player interactions. By integrating badminton-specific cues, strong performance and high interpretability are achieved.
Since badminton strokes exhibit clear temporal structure and strong dependencies on adjacent actions, incorporating categorical context from preceding and subsequent strokes is expected to better capture inter-stroke relationships.
Accordingly, stroke-type information from both the preceding and subsequent strokes is introduced as auxiliary categorical supervision to enhance temporal dependency modeling during training.
Formally, for the $i$-th sample in a batch, the preceding and subsequent stroke categories are defined as
\begin{equation}
\label{eq:pre_next_stroke}
\begin{aligned}
s_i^{\text{pre}} &\in \{0, 1, \dots, 30\}, \\
s_i^{\text{next}} &\in \{0, 1, \dots, 30\}, \quad i = 1, \dots, B,
\end{aligned}
\end{equation}
where $B$ denotes the batch size, and $30$ represents the total number of stroke categories, each corresponding to a specific type of badminton stroke.

Furthermore, the preceding and subsequent stroke categories are jointly considered to form a stroke-level temporal context representation:
\begin{equation}
\label{eq:stroke_context}
\mathbf{s} =
\bigl\{
\left( s_i^{\text{pre}},\, s_i^{\text{next}} \right)
\;\big|\;
i = 1, \dots, B
\bigr\}.
\end{equation}
This formulation allows explicit modeling of temporal dependencies between adjacent strokes, thereby facilitating a more comprehensive understanding of stroke sequence dynamics.
Each badminton video is represented as a sequence of frames, from which multimodal information—including player skeleton motion, shuttlecock trajectories, and on-court player positions—is extracted and jointly modeled. A video sequence $\mathbf{V}$ consisting of $F$ frames is denoted as
$\mathbf{V} = [\mathbf{v}_1, \mathbf{v}_2, \dots, \mathbf{v}_F]$.
For each frame, the skeleton motion of both players ($m = 1, 2$) is represented using joint and bone features. The joint representation of player $m$ at frame $f$ is defined as
$\mathbf{X}_f^{(m)} = \{ \mathbf{x}_{f,n}^{(m,t)} \}_{n=1,\ t=1}^{N,\ T} \in \mathbb{R}^{N \times T \times 2}$,
where $N$ denotes the number of joints, $T$ indicates the temporal sequence length, and each joint is described by its 2D image-plane coordinates.
Bone features are defined as relative displacement vectors between predefined pairs of connected joints that correspond to the human skeletal structure. Accordingly, the bone representation of player $m$ at frame $f$ is given by
$\mathbf{K}_f^{(m)} = \{ \mathbf{k}_{f,k}^{(m,t)} \}_{k=1,\ t=1}^{K,\ T} \in \mathbb{R}^{K \times T \times 2}$,
where $K$ denotes the number of bones, and each bone vector encodes the 2D displacement between a connected joint pair. The joint and bone features are concatenated along the joint--bone dimension to form the final skeleton representation:$\mathbf{JB}_f^{(m)} = \mathbf{X}_f^{(m)} \,\Vert\, \mathbf{K}_f^{(m)} \in \mathbb{R}^{(N+K) \times T \times 2}$.
In addition to human motion cues, shuttlecock trajectories are extracted using TrackNetV3~\cite{tracknetv3}. The shuttlecock position at frame $f$ is represented as
$\mathbf{C}_f = \{ c_{f,x}^{(t)}, c_{f,y}^{(t)} \}_{t=1}^{T} \in \mathbb{R}^{T \times 2}$,
where each element corresponds to the 2D image-plane coordinates of the shuttlecock at temporal step $t$.
Player positions on the court are estimated based on court line detection. Court lines are extracted using either the method proposed in MonoTrack~\cite{MonoTrack} or a deep learning-based approach, TennisCourtDetector~\cite{TennisCourtDetection}. The position of player $m$ at frame $f$ is defined as
$\mathbf{P}_f^{(m)} = \{ p_{f,x}^{(m,t)}, p_{f,y}^{(m,t)} \}_{t=1}^{T} \in \mathbb{R}^{T \times 2}$.
By integrating skeleton motion, shuttlecock trajectories, and on-court positional information into a unified representation, the resulting input enables simultaneous modeling of individual temporal dynamics and inter-player interactions. This joint representation is subsequently used as input to the TemPose~\cite{TemPose} framework for learning.
 

\subsection{TemPose-TF-ASF}
\label{sec:two_stage_stroke_fusion}
TemPose-TF-ASF extends TemPose-TF~\cite{TemPose} by incorporating predicted class information from both preceding and subsequent strokes as semantic priors.
This bidirectional temporal context enables context-aware refinement of target stroke predictions, improving classification accuracy and temporal consistency.

\subsubsection{TemPose-TF~\cite{TemPose}}
\label{sec:TemPose-TF}
\begin{figure*}[htbp]
    \centering
        \includegraphics[width=1.0\linewidth]{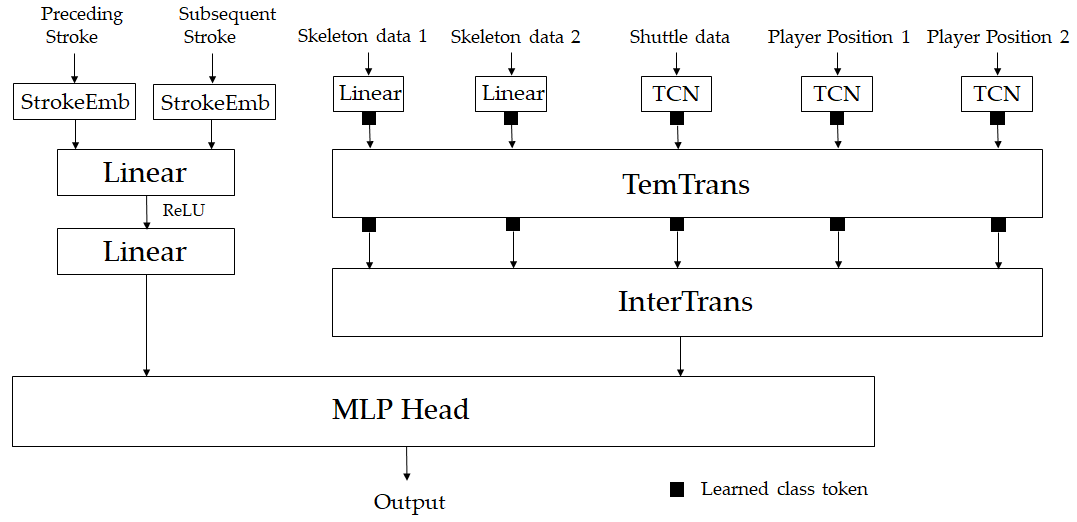}

    \caption{
Architecture of TemPose-TF-ASF
TemPose-TF-ASF incorporates semantic information from both preceding and subsequent strokes, enabling context-aware refinement of the target stroke prediction.
}  
    \label{fig:Architecture of TemPose-TF-ASF-a}
\end{figure*}

TemPose adopts a factorized Transformer encoder that decouples temporal modeling from inter-person interaction modeling, enabling effective handling of multi-person action sequences~\cite{TemPose}.
The architecture consists of a Temporal Transformer Layer and an Interaction Transformer Layer.
In the Temporal Transformer Layer, temporal dynamics are modeled independently for each skeleton sequence, capturing action evolution at the individual level.
This design supports parallel processing of up to $M$ individuals and produces a temporal class token as a compact semantic representation for each person.
After encoding the temporal features of all individuals, the Interaction Transformer Layer treats the resulting temporal class tokens as interaction units. By introducing an interaction class token and interaction embeddings, the model explicitly captures inter-person relationships. The interaction features are then processed by subsequent Transformer layers and an multilayer perceptron (MLP) head to predict the overall action category.
Based on these architectural properties, TemPose-TF is adopted as the backbone of TemPose-TF-ASF, as illustrated in Fig.~\ref{fig:Architecture of TemPose-TF-ASF-a}. In the Temporal Fusion configuration, player position and shuttlecock position information are first processed independently through dedicated Temporal Convolutional Network (TCN) modules to extract temporal features \cite{TCN2}. Each TCN consists of two one-dimensional convolutional layers with dilation rates of 1 and 3, respectively, using a kernel size of 5, and a stride of 1.
Following TCN processing, positional features are projected into the same embedding space as the skeleton tokens and appended to the input of the Temporal Transformer Layer as additional person tokens. This design allows skeletal motion and on-court positional information to be jointly considered during temporal modeling. Subsequently, all temporal class tokens are aggregated by the Interaction Transformer Layer, and the final action category is predicted via an MLP head.

\subsubsection{Two-Stage Contextual Refinement (TSCR) via Estimated Target Stroke Feedback}
\label{sec:two_stage}
\begin{figure*}[tp]
    \centering

        \includegraphics[width=\linewidth]{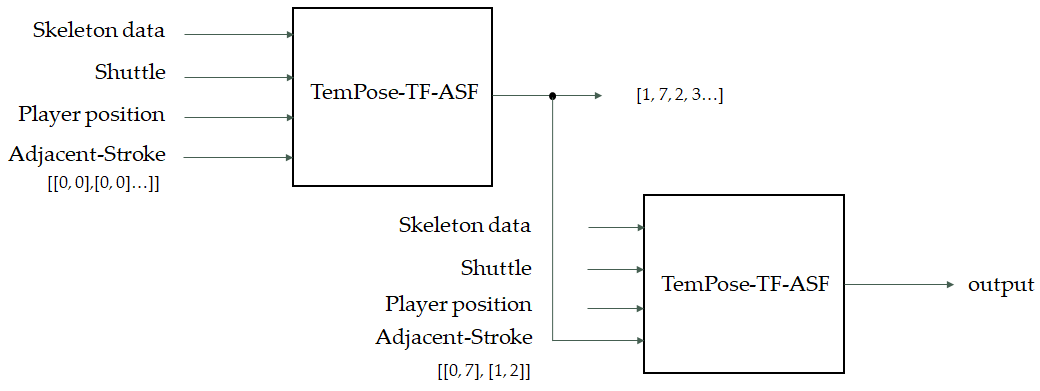}
        
\caption{
 TSCR Pipeline of TemPose-TF-ASF
For TemPose-TF-ASF, Adjacent-Stroke inputs are initialized as zero vectors in the first stage, reducing the model to the original TemPose-TF~\cite{TemPose} for context-free stroke prediction. The predicted stroke labels within each batch are then aggregated and reformulated as estimated Adjacent-Stroke inputs. In the second stage, these estimated labels are fed back to refine stroke predictions with explicit contextual guidance.
}
      \label{fig:Two-Stage Contextual Refinement Pipeline of TemPose-TF-ASF-a}
    \end{figure*}

TemPose-TF-ASF adopts a TSCR training and inference strategy,
as illustrated in Fig.~\ref{fig:Two-Stage Contextual Refinement Pipeline of TemPose-TF-ASF-a}. In addition to the original inputs, auxiliary categorical information from the preceding and subsequent strokes is incorporated to further enhance stroke recognition performance.
In the first stage, the preceding and subsequent stroke inputs are set to zero, indicating the absence of neighboring temporal context. The model therefore relies solely on the current stroke segment, making it functionally equivalent to the original TemPose-TF~\cite{TemPose}, and is trained to predict stroke labels using only current-stroke features.
The estimated target stroke sequence obtained from the first stage is then reorganized into preceding and subsequent stroke categories, as defined in Eqs.~(\ref{eq:pre_next_stroke}) and~(\ref{eq:stroke_context}).
These stroke-level temporal context representations are subsequently fed into the second stage, where refined predictions are produced by explicitly leveraging information from adjacent strokes.

\subsubsection{Stroke Embedding and Bidirectional Context Fusion}
\label{sec:stroke_fusion_detail}
\begin{figure}[tp]
    \centering
    \includegraphics[width=1.0\linewidth]{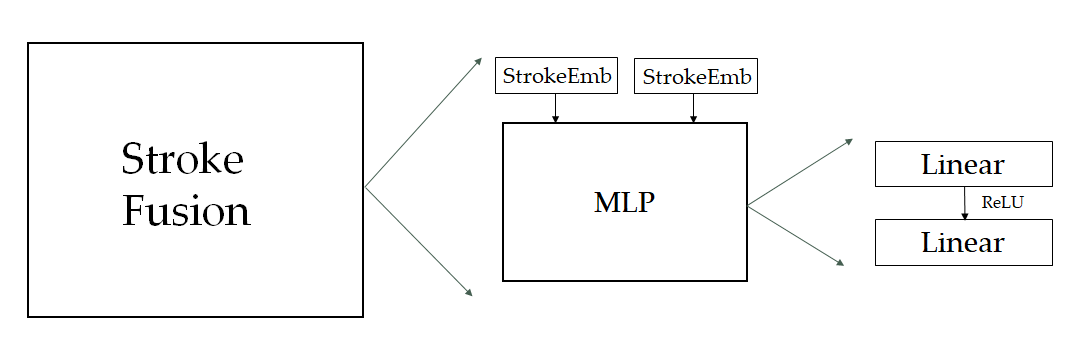}
    \caption{Architecture of Stroke Fusion}
    \label{fig:Stroke Fusion}
\end{figure}

The preceding and subsequent stroke information fed into the model is processed by the Stroke Fusion module, which is designed to integrate and transform stroke-level temporal contextual information. The architecture of the Stroke Fusion module is illustrated in Fig.~\ref{fig:Stroke Fusion}. The overall procedure consists of three main steps: stroke embedding, feature concatenation, and MLP-based fusion.
First, categorical stroke labels are projected into a continuous embedding space.
For the $i$-th sample in a batch, the embeddings of the preceding and subsequent strokes
are defined as
\begin{equation}
\begin{aligned}
\mathbf{e}_i^{\text{pre}}  &= \mathrm{StrokeEmb}\!\left( s_i^{\text{pre}} \right), \\
\mathbf{e}_i^{\text{next}} &= \mathrm{StrokeEmb}\!\left( s_i^{\text{next}} \right).
\end{aligned}
\end{equation}

where $\mathrm{StrokeEmb}(\cdot)$ denotes a learnable embedding layer that maps each stroke category to a $D$-dimensional vector representation.
Next, the embeddings of the preceding and subsequent strokes are concatenated along the feature dimension to form a joint representation:
\begin{equation}
\mathbf{e}_i^{\text{cat}} =
\left[ \mathbf{e}_i^{\text{pre}} \,\Vert\, \mathbf{e}_i^{\text{next}} \right]
\in \mathbb{R}^{2D}.
\end{equation}
Finally, the concatenated feature is passed through an MLP composed of two fully connected layers with a ReLU nonlinearity in between, producing the final fused stroke representation:
\begin{equation}
\mathbf{f}_i = \phi\!\left( \mathbf{e}_i^{\text{cat}} \right)
\in \mathbb{R}^{H},
\end{equation} where $\phi(\cdot)$ denotes the MLP-based fusion function, and $H$ represents the dimensionality of the hidden feature space.
The resulting fused representation $\mathbf{f}_i$ jointly encodes the temporal context of both the preceding and subsequent strokes and is provided as an additional input to facilitate more accurate and temporally consistent stroke recognition.

\subsection{TemPose-TF-BiLP and TemPose-TF-BiTP}
\label{sec:TemPose-TF-BiLP}
TemPose-TF-BiLP (Bidirectional LSTM Predictor) and TemPose-TF-BiTP (Bidirectional Transformer Predictor) perform preliminary stroke classification using forward and backward predictors to model bidirectional temporal stroke sequences. Forward and backward LSTM or Transformer predictors capture complementary temporal dependencies from past and future contexts, and their stroke predictions are fused by the Stroke Fusion module for refinement. This design incorporates bidirectional temporal cues while avoiding reliance on neighboring stroke ground-truth information during inference.

\subsubsection{Forward and Backward LSTM Predictor (LP)}
\label{sec:LSTMPredictor}
\begin{figure*}[tb]
    \centering
    \begin{subfigure}{0.48\linewidth}
        \centering
        \includegraphics[width=0.8\linewidth]{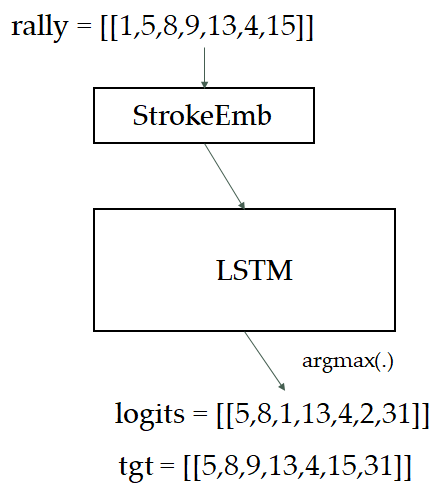}
        
        \caption{Architecture of Forward LP
        The Forward LP processes the rally sequence in chronological order under a sequence-to-sequence paradigm~\cite{sequence-to-sequence}, predicting the next-stroke class at each time step. It outputs class logits supervised by ground-truth labels via a cross-entropy loss, and the resulting predictions provide forward contextual cues for the main model.
        }
        \label{fig:Forward LSTMPredictor-a}
    \end{subfigure}
    \hfill
    \begin{subfigure}{0.48\linewidth}
        \centering
        \includegraphics[width=0.8\linewidth]{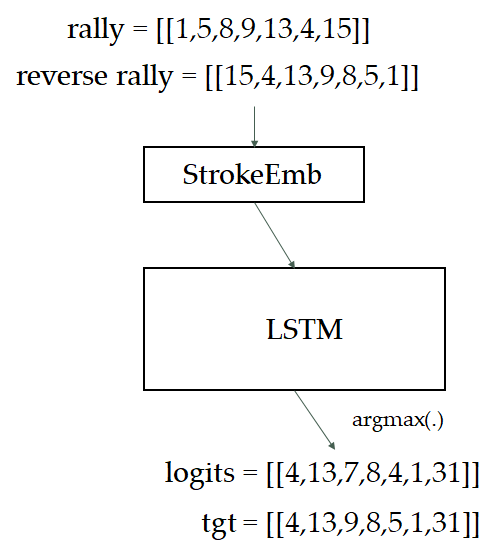}
       
        \caption{Architecture of Backward LP
        The Backward LP processes a temporally reversed rally sequence under a sequence-to-sequence paradigm~\cite{sequence-to-sequence}, predicting next-stroke classes in the reversed temporal direction. The resulting logits are supervised by cross-entropy loss with corresponding target labels, enabling effective modeling of backward temporal dependencies.
        }
        \label{fig:Backward LSTMPredictor-b}
    \end{subfigure}

    \caption{
       Architecture of the LSTM Predictor (LP).}
    \label{fig:LSTMPredictor}
\end{figure*}

Given a stroke class sequence of length $L$, denoted as $\{s_1, s_2, \dots, s_L\}$, discrete stroke class indices are first transformed into a continuous embedding space. Specifically, each stroke class $s_t$ is mapped to a $D$-dimensional vector through a learnable stroke embedding function $\mathrm{StrokeEmb}(\cdot)$, facilitating subsequent temporal feature learning.
As illustrated in Fig.~\ref{fig:Forward LSTMPredictor-a}, the Forward LP processes the embedded stroke sequence in its original temporal order and employs a multi-layer LSTM network to model long-range temporal dependencies among strokes. At each time step $t$, the LSTM produces a hidden state that summarizes historical stroke information up to that time. The hidden state at time step $t$ primarily captures temporal context from the first stroke to the $t$-th stroke, making it suitable for modeling the influence of past strokes on the current stroke prediction.
To additionally exploit semantic cues provided by subsequent strokes, a Backward LP is constructed by processing the temporally reversed stroke sequence. Through backward temporal modeling, complementary contextual information from future strokes is captured, enhancing the representation of the target stroke. In practice, the Backward LP shares the same network architecture as the Forward LP, as illustrated in Fig.~\ref{fig:Backward LSTMPredictor-b}. During feature extraction, temporal indices are re-aligned based on the actual sequence length to ensure that the extracted hidden states correctly correspond to the target stroke positions in the original temporal order.

\subsubsection{Forward and Backward Transformer Predictor (TP)}
\label{sec:TransformerPredictor}
\begin{figure*}[t]
    \centering
    \begin{subfigure}{0.48\linewidth}
        \centering
        \includegraphics[width=\linewidth]{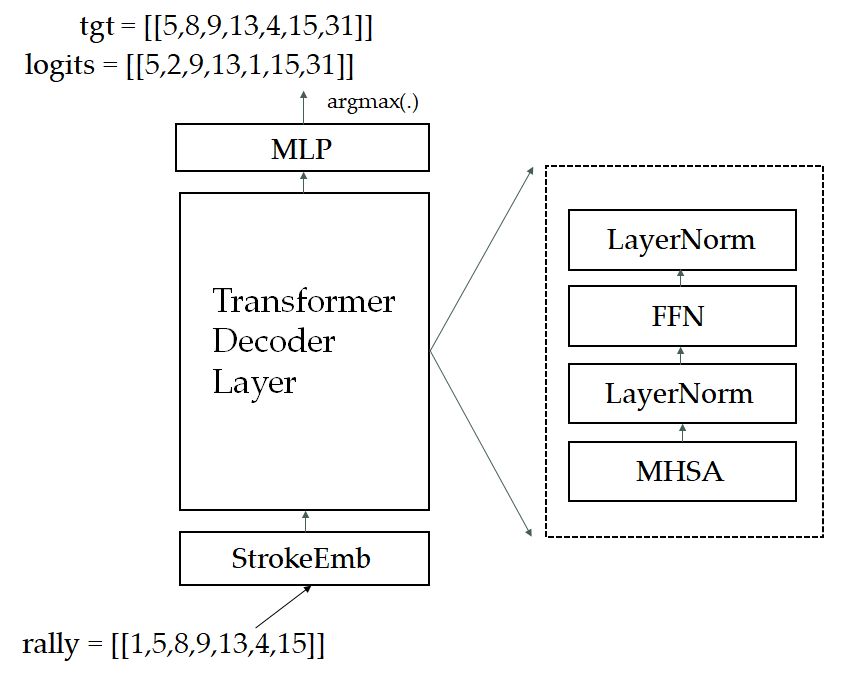}
        \caption{Architecture of Forward TP
        The Forward TP processes the rally sequence in chronological order under a sequence-to-sequence paradigm~\cite{sequence-to-sequence}, predicting the next-stroke class at each time step. It outputs class logits supervised by cross-entropy loss with ground-truth labels, and the resulting predictions provide forward contextual cues for the main model.
        }
        \label{fig:Forward TransformerPredictor-a}
    \end{subfigure}
    \hfill
    \begin{subfigure}{0.48\linewidth}
        \centering
        \includegraphics[width=\linewidth]{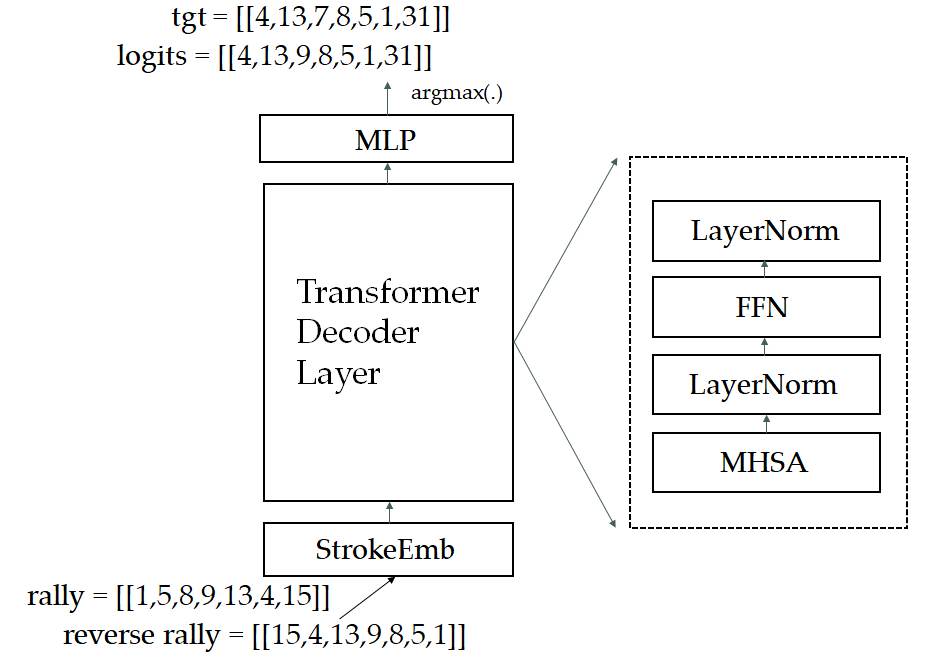}
        \caption{Architecture of Backward TP
      The Backward TP operates on a temporally reversed rally sequence, in which the original stroke order is inverted. A sequence-to-sequence learning framework is similarly employed to predict the next stroke class in the reversed temporal direction, with logits optimized against the corresponding target labels (tgt). This design allows the model to capture backward temporal dependencies within the rally sequence.  
        }
        \label{fig:Backward TransformerPredictor-b}
    \end{subfigure}
 \caption{
Architecture of the Transformer Predictor (TP).
    }   
    \label{fig:TransformerPredictor}
\end{figure*}

To model long-range temporal dependencies in stroke sequences, Forward and Backward Transformer Predictors (TPs) are introduced, replacing recurrent architectures with self-attention mechanisms for temporal modeling. As shown in Fig.~\ref{fig:Forward TransformerPredictor-a} and  Fig.~\ref{fig:Backward TransformerPredictor-b}, the Forward and Backward TPs share the same network architecture and differ only in the temporal ordering of the input stroke sequences and the index alignment strategy adopted during feature extraction.
In the Forward TP, the stroke sequence is processed in its original temporal order using multiple Transformer decoder layers. To prevent access to future information during prediction, a causal mask is applied within the self-attention mechanism, restricting each time step to attend only to itself and previous strokes.
Under this setting, the hidden representation produced at time step $t$ primarily reflects accumulated temporal context from the first stroke to the $t$-th stroke, effectively modeling the influence of past strokes on the current stroke prediction.
To exploit semantic cues from future strokes, a Backward Transformer Predictor (TP) is employed. Its input consists of the temporally reversed stroke sequence, enabling backward modeling from future to past. This reverse modeling captures complementary temporal information that may not be fully modeled by forward processing alone.
During feature extraction, temporal indices are computed based on the sequence length and target stroke position to ensure correct temporal alignment. Features from the Forward TP are extracted at the target stroke time step, whereas features from the Backward TP are re-aligned according to the sequence length to map reverse-modeled representations back to the original temporal order.

\subsubsection{Bidirectional Stroke Prediction Framework}
\label{sec:target-stroke}
This subsection presents the bidirectional stroke prediction for TemPose-TF-BiLP and TemPose-TF-BiTP.
First, a forward predictor is trained on the original stroke order to model temporal dependencies, using either a Forward LP or a Forward TP to predict the target stroke from preceding strokes.
Next, a backward predictor is trained on the temporally reversed sequence, where a Backward LP or Backward TP captures complementary temporal information from subsequent strokes through backward modeling.
Finally, the forward and backward predictors are jointly utilized to generate estimated target stroke predictions. These predictions are incorporated into a TSCR framework, where estimated stroke labels are successively fed into TemPose-TF-BiLP or TemPose-TF-BiTP and integrated through the Stroke Fusion module to progressively refine the final stroke classification. 


\section{Experiment}
\label{sec:experiments}

\begin{table*}[tp]
    \centering
    \caption{Overview of model architectures and temporal auxiliary input settings for TemPose TemPose and its variants, together with SOTA methods, used in the stroke recognition task.}
  \label{tab:comparison_strategy1-1}
    \begin{tabular}{l|c|ccccc}\toprule
        
        Model   & Param& Modality& pos + shuttle&Pre stroke &next stroke &target stroke \\\midrule
 BlockGCN~\cite{BlockGCN}&1.69M& J-only& \texttimes& \texttimes& \texttimes&\texttimes\\
 SkateFormer~\cite{SkateFormer}&2.38M& J-only& \texttimes& \texttimes& \texttimes&\texttimes\\
 TemPose-V~\cite{TemPose}&1.62M& JnB& \texttimes& \texttimes& \texttimes&\texttimes\\
 TemPose-TF~\cite{TemPose}& 1.71M& JnB& \checkmark& \texttimes& \texttimes& \texttimes\\
 BST-CG-AP~\cite{chang2025bst}& 1.88M& JnB& \checkmark& \texttimes& \texttimes& \texttimes\\
 BlockGCN-ASF&2.09M& J-only& \texttimes& \checkmark& \checkmark&\texttimes\\
 SkateFormer-ASF&2.71M& J-only& \texttimes& \checkmark& \checkmark&\texttimes\\
 TemPose-V-ASF&1.86M& JnB& \texttimes& \checkmark& \checkmark&\texttimes\\
 BST-CG-AP-ASF& 2.12M& JnB& \checkmark& \checkmark& \checkmark& \texttimes\\
 TemPose-TF-ASF& 1.95M& JnB& \checkmark& \checkmark& \checkmark& \texttimes\\
\midrule
        TemPose-TF~\cite{TemPose}   &1.71M& JnB& \checkmark& \texttimes& \texttimes&\texttimes\\
        TemPose-TF-ASF  &1.95M& JnB& \checkmark& \checkmark& \checkmark&\texttimes\\
        TemPose-TF-PSF  &1.92M& JnB& \checkmark& \checkmark& \texttimes&\texttimes\\
 TemPose-TF-TSF  &1.92M& JnB& \checkmark& \texttimes& \texttimes&\checkmark\\
 TemPose-TF-DualSF  &2.02M& JnB& \checkmark& \checkmark& \texttimes&\checkmark\\
 TemPose-TF-NSF  &1.92M& JnB& \checkmark& \texttimes& \checkmark&\texttimes\\
 TemPose-TF-TriSF  &2.16M& JnB& \checkmark& \checkmark& \checkmark&\checkmark\\ \bottomrule
    \end{tabular}
\end{table*}

\begin{table*}[tp]
    \centering
      \caption{Quantitative comparison of TemPose \cite{TemPose} variants with state-of-the-art methods on the stroke recognition task. Under different temporal auxiliary supervision settings, results are reported in terms of Acc, Macro-F1, and Acc-2.}
  \label{tab:comparison_strategy1-2}
    \begin{tabular}{l|c|ccc}\toprule
        
        Model   & Param& Acc & Macro-F1 & Acc-2 \\\midrule
 BlockGCN~\cite{BlockGCN}&1.69M& 0.793& 0.657&0.906\\
 SkateFormer~\cite{SkateFormer}& 2.38M& 0.808& 0.665&0.897\\
 TemPose-V~\cite{TemPose}&1.62M& 0.798& 0.658&0.918\\
 TemPose-TF~\cite{TemPose}& 1.71M& 0.835& 0.742&0.927\\
 BST-CG-AP~\cite{chang2025bst}& 1.88M& 0.838& 0.723&0.942\\
 BlockGCN-ASF&2.09M& 0.809& 0.673&0.925\\
 SkateFormer-ASF&2.71M& 0.811& 0.673&0.901\\
 TemPose-V-ASF&1.86M& 0.814& 0.701&0.921\\
 BST-CG-AP-ASF& 2.12M& 0.843& 0.729&0.944\\
 TemPose-TF-ASF& 1.95M& \textbf{0.854}& \textbf{0.761}&\textbf{0.945}\\
\midrule
        TemPose-TF~\cite{TemPose}   &1.71M& 0.835& 0.742& 0.927\\
        TemPose-TF-ASF  &1.95M& \textbf{0.854}& \textbf{0.761}& 0.945\\
        TemPose-TF-PSF  &1.92M& 0.853& 0.746& \textbf{0.948}\\
 TemPose-TF-TSF  &1.92M& 0.848& 0.745&0.940\\
 TemPose-TF-DualSF  &2.02M& 0.847& 0.742&0.942\\
 TemPose-TF-NSF  &1.92M& 0.850& 0.751&0.941\\
 TemPose-TF-TriSF  &2.16M& 0.847& 0.738&0.944\\ \bottomrule
    \end{tabular}
\end{table*}
        
\begin{table*}[htbp]
    \centering
        \caption{Comparison of TemPose~\cite{TemPose} and its extended variants incorporating predicted stroke-type information for the target stroke.}
   \label{tab:comparison_strategy2}
    \begin{tabular}{l|c|ccc}
        \toprule
        Model &  Param&Acc & Macro-F1 & Acc-2 \\
        \midrule
        TemPose-TF~\cite{TemPose} &  1.71M&0.835& 0.742& 0.927\\
        TemPose-TF-TSF&  1.92M&0.848& \textbf{0.745}& 0.940\\
 TemPose-TF-BiLP&  6.09M&\textbf{0.850}& 0.737&\textbf{0.945}\\
 TemPose-TF-BiTP&  5.37M&0.848& 0.741&0.939\\ \bottomrule
    \end{tabular}
\end{table*}


\begin{table*}[tp]
    \centering
        \caption{Performance comparison of two-stage versus three-stage target-stroke prediction extensions for TemPose~\cite{TemPose}.}
\label{tab:comparison_strategy3}
    \begin{tabular}{l|c|ccc}
        \toprule
        Model &  Param&Acc & Macro-F1 & Acc-2 \\\midrule
        TemPose-TF-SiLP&  4.99M&\textbf{0.847}& 0.734& \textbf{0.943}\\
 TemPose-TF-SiTP&  4.57M&0.843& 0.742&0.935\\
 TemPose-TF-PSSiLP&  5.09M&0.843& 0.736&0.939\\
 TemPose-TFwith-PSSiTP&  4.78M&0.841& \textbf{0.747}&0.935\\ \bottomrule
    \end{tabular}
\end{table*}
\subsection{Dataset}
\label{subsec:dataset}
ShuttleSet~\cite{ShuttleSet} is currently the largest publicly available badminton video dataset. It consists of 44 singles matches recorded at the top competitive level between 2018 and 2021, covering 27 high-ranking male and female singles players. In total, ShuttleSet includes 104 games, comprising 3,685 rallies and 36,492 shots.
The original ShuttleSet ~\cite{ShuttleSet} contains 19 distinct stroke categories, including a "none" class.\cite{chang2025bst,ShuttleSet}. To ensure annotation quality, erroneous or problematic samples were removed through a data cleaning process. After cleaning, 40 matches were retained and split into 30 matches for training, 5 for validation, and 5 for testing, resulting in a total of 33,429 annotated strokes.
In addition, two stroke categories with fewer than 50 instances in the entire dataset, namely “push” and “lob”, were merged into a single category labeled “push lob”,  while “wrist smash” was merged into the “smash” category. To further distinguish between upper-and lower-court players, all stroke categories except ``none'' were duplicated according to player court position. After these adjustments, the final classification setup consists of 31 stroke categories.
\subsection{Performance Comparison of Stroke Fusion Strategies }
The proposed TemPose-TF-ASF model enhances badminton stroke prediction by integrating stroke-type information from both preceding and subsequent strokes. As shown in \cref{tab:comparison_strategy1-1} and \cref{tab:comparison_strategy1-2}  , TemPose-TF-ASF achieves substantial performance improvements over the baseline TemPose-TF~\cite{TemPose}, demonstrating the effectiveness of incorporating bidirectional temporal stroke context for stroke recognition.
Furthermore, comparisons with several state-of-the-art (SOTA) methods, including BlockGCN~\cite{BlockGCN}, SkateFormer~\cite{SkateFormer}, and BST-CG-AP~\cite{chang2025bst}, were conducted. The proposed ASF module was integrated into these architectures, resulting in BlockGCN-ASF, SkateFormer-ASF, and BST-CG-AP-ASF. All augmented models consistently outperform their original versions. These results indicate that the temporal stroke semantic information introduced by ASF effectively complements limitations in existing temporal context modeling. The consistent gains across different architectures further demonstrate that ASF is backbone-agnostic and exhibits strong transferability and generalization.
In addition to the full ASF design, several variant models incorporating only partial stroke-type information were evaluated. These variants include TemPose-TF-PSF (Pre-Stroke Fusion), which incorporates only preceding strokes; TemPose-TF-NSF (Next-Stroke Fusion), which incorporates only subsequent strokes; TemPose-TF-TSF (Target-Stroke Fusion), which uses only the estimated target stroke label embedding; TemPose-TF-DualSF (Dual-Stroke Fusion), which combines preceding and estimated target strokes; and TemPose-TF-TriSF (Triple-Stroke Fusion), which integrates preceding, subsequent, and estimated target strokes. Quantitative results for these variants are summarized in \cref{tab:comparison_strategy1-1} and \cref{tab:comparison_strategy1-2}.
Experimental results indicate that TemPose-TF-ASF achieves the best overall performance among all evaluated variants in terms of both Accuracy and Macro-F1. This observation suggests that jointly modeling bidirectional temporal context provides more complete and informative temporal cues for stroke classification. While variants using estimated target stroke label embeddings achieve some gains, the overall improvement is limited. TemPose-TF-TriSF does not outperform TemPose-TF-ASF, suggesting that excessive semantic input can introduce redundancy or noise and reduce temporal context modeling effectiveness. These results confirm that selectively incorporating temporally and semantically relevant stroke information is more effective than indiscriminately increasing fused features for improving badminton stroke recognition.
\subsection{Evaluation of Alternative Architectures for Target Stroke Integration }
 \cref{tab:comparison_strategy2} presents a comparison of three strategies for integrating the predicted target stroke type into the TemPose \cite{TemPose} framework. The evaluated approaches include TemPose-TF-TSF (Target-Stroke Fusion), which embeds the stroke type predicted in the first stage, and TemPose-TF-BiLP and TemPose-TF-BiTP, which use forward and backward rally sequences to predict the target stroke and extract contextual features from predictor hidden states. The results indicate that, although TemPose-TF-TSF relies on estimated target labels rather than ground-truth annotations, the relatively high semantic clarity and low noise of the embedded labels enable this approach to achieve the highest Macro-F1 score. This observation suggests that estimated stroke labels can still provide effective class-level semantic guidance when incorporated in a controlled and lightweight manner.
 In contrast, TemPose-TF-BiLP achieves the highest Accuracy, indicating that LSTM-based predictors provide more stable modeling of long-range temporal dependencies and reliable contextual support.
TemPose-TF-BiTP has stronger representational capacity, but its training is less stable on the limited dataset, resulting in lower Accuracy and Macro-F1 scores.
Overall, these results highlight a trade-off between class-balanced performance and overall accuracy. Embedding estimated stroke labels improves Macro-F1, while LSTM-based predictors better enhance Accuracy. Under the current dataset scale, the Transformer Decoder-based predictor shows no clear advantage.
\subsection{Comparison Between Two-Stage and Three-Stage Sequential Inference Architectures}
 \cref{tab:comparison_strategy3} compares two-stage and three-stage sequential inference architectures. In the two-stage setting, TemPose-TF-SiLP (TemPose-TF with Single LSTM Predictor) and TemPose-TF-SiTP (TemPose-TF with Single Transformer Predictor) employ only forward LP and TP, respectively, to estimate the target stroke class, and the predicted class is integrated into the main model to facilitate more refined stroke inference. In contrast, the three-stage architectures, including TemPose-TF-PSSiLP (TemPose-TF with Pre-Stroke Single LSTM Predictor) and TemPose-TF-PSSiTP (TemPose-TF with Pre-Stroke Single Transformer Predictor), transform the predictions obtained from the same overall framework as TemPose-TF-SiLP and TemPose-TF-SiTP into preceding-stroke information, which is then fed into a third stage to produce the final stroke classification.
 Experimental results show that two-stage architectures achieve higher Accuracy and Top-2 Accuracy than three-stage counterparts, with TemPose-TF-SiLP being the most stable among them.
Three-stage architectures improve Macro-F1, with TemPose-TF-PSSiTP reaching 0.747, the highest among all models. This indicates that using the predicted previous stroke class enhances discrimination for minority categories and balances performance across classes.
However, the three-stage pipeline relies on an additional prediction signal, making it more prone to error propagation and reducing overall Accuracy.
In summary, two-stage architectures are better for overall accuracy and inference stability, while three-stage architectures are preferable for improving minority-class performance in imbalanced datasets.
\section{Conclusion}
\label{sec:conclusion}
TemPose-TF-ASF, a context-aware extension of TemPose~\cite{TemPose}, is proposed to enhance temporal context modeling for badminton stroke recognition by incorporating stroke-type information from both preceding and subsequent strokes. Experimental results show that bidirectional stroke context consistently and substantially improves performance over the baseline. The TSCR strategy further enhances accuracy while mitigating the training–inference mismatch, without requiring ground-truth annotations of future strokes. Performance gains are also observed when the ASF module is integrated into multiple state-of-the-art temporal action recognition frameworks, demonstrating strong transferability and generalization. Overall, jointly exploiting temporally and semantically relevant stroke context with a staged learning strategy is shown to be an effective and practical approach for improving badminton stroke recognition.
{
    \small
    \bibliographystyle{ieeenat_fullname}
    \bibliography{main}
}


\end{document}